%% file: dsscore2_main.tex
\documentclass[preprint,12pt,authoryear]{elsarticle}

\usepackage{amssymb}
\usepackage{amsmath}
\usepackage{etoolbox}
\usepackage{color}
\journal{Artificial Intelligence in Medicine}

\include{0_colortodoins}

\newcommand{\kmddss}{
\texttt{TreatmentRecommender}
}
\begin{document}

\include{0_my_frontmatter}


\section{Introduction}
\label{sec:intro}
\input{1_introduction}

\section{Related Work}
\label{sec:relwork}
\input{2_relwork}

\section{Materials}
\label{sec:materials}
\input{3_materials}

\section{Our \kmddss{}}
\label{sec:ourmethod}
\input{4_ourDSScore}

\section{Experimental Results}
\label{sec:experiments}
\input{5_results}

\section{Discussion}
\label{sec:discussion}
\input{6_discussion}

\section{Summary and Outlook}
\label{sec:closing}
\input{7_conclusion}

\section*{Acknowledgements}
Work of the authors VU, CP, MSch, UN, RC and EV was partially funded through the European Union’s Horizon 2020 Research and Innovation Programme under grant agreement number 848261 "Unification of treatments and Interventions for Tinnitus patients" (UNITI).

\section*{Author Contributions}
Vishnu Unnikrishnan (VU): design of data preparation, learning methods and clinical decision support system back-end; experimental evaluation; drafted the manuscript. /

Clara Puga (CP): contributed to data preparation, learning methods and clinical decision support system back-end./

Miro Schleicher (MSch): contributed to data preparation, learning methods and clinical decision support system back-end./

Uli Niemann (UN): contributed to the design of the analytics workflow./

Berthold Langguth (BL): contributed to RCT design and to model validation against the RCT; participated in RCT data collection; provided critical feedback on clinical decision support system design. /

Stefan Schoisswohl (SSch): contributed to RCT design and to model validation against the RCT; participated in RCT data collection; provided critical feedback on clinical decision support system design. / 

Birgit Mazurek (BM): contributed to RCT design and to model validation against the RCT; participated in RCT data collection; provided critical feedback on clinical decision support system design./

Jose Antonio López-Escámez (JALE): contributed to RCT design and to model validation against the RCT; participated in RCT data collection; provided critical feedback on clinical decision support system design./

Rilana Cima (RC): contributed to RCT design and to model validation against the RCT; participated in RCT data collection; provided critical feedback on clinical decision support system design./

Dimitris Kikidis (DK): contributed to RCT design and to model validation against the RCT; participated in RCT data collection; provided critical feedback on clinical decision support system design. /

Eleftheria Vellidou (EV) contributed to RCT design and to model validation against the RCT; provided critical feedback on clinical decision support system design. /

Winfried Schlee (WS): contributed to RCT design and to model validation against the RCT; participated in RCT data collection; provided critical feedback to data preparation design; provided critical feedback on clinical decision support system design; provided critical feedback to manuscript./

Rüdiger Pryss (RP): contributed to RCT design and to model validation against the RCT; contributed to data cleaning and preparation; provided critical feedback on clinical decision support system design./

Myra Spiliopoulou (MSp): design of data preparation, learning methods and clinical decision support system back-end; contributed to manuscript design and writing. /

The authors read and approved the final manuscript.
%\appendix
%\section{Example Appendix Section}
%\label{app1}

%%       \citet{<label>} ==> Lamport (1994)
%%       \citep{<label>} ==> (Lamport, 1994)
%%
%Example citation, See \citet{lamport94}.

\bibliographystyle{elsarticle-harv} 
\bibliography{KMD_2019onwards.bib,%
medicalDecisionSupport_AIIM.bib,%
medicalDecisionSupport_scholar.bib,%
otherBIB.bib,%
questionnaires_BIB.bib
}



%\begin{thebibliography}{00}
%
%%% For authoryear reference style
%%% \bibitem[Author(year)]{label}
%%% Text of bibliographic item
%
%\bibitem[Lamport(1994)]{lamport94}
%  \textit{\LaTeX: a document preparation system},
%  Addison Wesley, Massachusetts,
%\end{thebibliography}
\end{document}

%% file: 0_colortodoins.tex
\usepackage[textsize=small,backgroundcolor=green]{todonotes}
\usepackage{url}

\definecolor{MyraColor}{RGB}{00,130,60}
\definecolor{TillColor}{RGB}{00,00,190}

\definecolor{MyraBGcolor}{RGB}{152,251,152} % palegreen
\definecolor{MyraFGcolor}{RGB}{0,100,0} % darkgreen

\definecolor{moreFGcolor}{RGB}{0,0,205} % mediumblue; feel free tochange
\definecolor{moreBGcolor}{RGB}{175,238,238} % paleturquoise; feel free to change

\definecolor{furtherFGcolor}{RGB}{120,180,00} %% olive green
\definecolor{furtherBGcolor}{RGB}{240,240,00} % yellow green

\definecolor{anyFGcolor}{RGB}{240,80,0} %orange
\definecolor{anyBGcolor}{RGB}{255,130,0} %orange

%% file: 0_my_frontmatter.tex
\begin{frontmatter}



\title{Training and Validating a Treatment 
%Outcome Predictor
Recommender with
%despite 
Partial Verification Evidence} %% Article title


\author[KMD]{Vishnu Unnikrishnan}
\author[KMD]{Clara Puga}
\author[KMD]{Miro Schleicher}
\author[BIBO]{Uli Niemann \footnote{Work done while with affiliation $^{a}$}}
\author[UHREG]{Berthold Langguth}
\author[UHREG,BUNDWM]{Stefan Schoisswohl}
\author[CHA]{Birgit Mazurek}
\author[KULEUVEN,ADEL,UMAAS]{Rilana Cima}
\author[GRA,KOL]{Jose Antonio Lopez-Escamez}
\author[HIPPOKRATION]{Dimitris Kikidis}
\author[NTUA]{Eleftheria Vellidou}
%\author[NTUA]{Konstantinos Bromis}
\author[UWUE]{Ruediger Pryss}
\author[UHREG,StGallen]{Winfried Schlee}
\author[KMD]{Myra Spiliopoulou} %% Author name

\affiliation[KMD]{organization={Knowledge Management \& Discovery Lab, Otto-von-Guericke-University Magdeburg},
%Department and Organization
%           city={Magdeburg},
%           postcode={}, 
            country={Germany}}
\affiliation[BIBO]{organization={University Library, Otto-von-Guericke-University Magdeburg},
%Department and Organization
%           city={Magdeburg},
%           postcode={}, 
            country={Germany}}
\affiliation[CHA]{organisation={Tinnitus Center, Charité - Universitätsmedizin Berlin, corporate member of Freie Universität Berlin und Humboldt Universität Berlin},
%city={Berlin},
country={Germany}}
\affiliation[GRA]{organisation={Department of Surgery, Division of Otolaryngology, Universidad de Granada},
%city={Granada},
country={Spain}}
\affiliation[KOL]{organisation={Meniere's Disease Neuroscience Research Program, Faculty of Medicine \& 
Health, School of Medical Sciences, The Kolling Institute, University of Sydney}, country={Australia}}

\affiliation[KULEUVEN]{organisation={Faculty of Psychology and Educational Sciences, Health Psychology, Katholieke Universiteit Leuven}, 
%city={Leuven, Flanders}, 
country={Belgium}}

\affiliation[ADEL]{organisation={Adelante, Centre of Expertise in Rehabilitation and Audiology}, city={Hoensbroek}, country={The Netherlands}}

\affiliation[UMAAS]{organisation={Faculty of Psychology and Neurosciences, Experimental Health Psychology, Maastricht University},
%city={Maastricht, Limburg},
country={The Netherlands}}

\affiliation[BUNDWM]{organisation={Department of Human Sciences, Institute of Psychology, Universitaet der Bundeswehr München}, city={Neubiberg}, country={Germany}}
\affiliation[HIPPOKRATION]{organization={First Department of Otorhinolaryngology, Head and Neck Surgery, National and Kapodistrian University of Athens, Hippokration General Hospital}, city={Athens}, country={Greece}}
\affiliation[NTUA]{organisation={Institute of Communications and Computer Systems (ICCS), National Technical University of Athens}, 
%city={Athens}, 
country={Greece}}
\affiliation[UHREG]{organisation={Department of Psychiatry and Psychotherapy, University of Regensburg}, 
%city={Regensburg},  
country={Germany}}
\affiliation[UWUE]{organisation={Institute of Clinical Epidemiology and Biometry, University of Würzburg}, 
%city={W\"urzburg},  
country={Germany}}
\affiliation[StGallen]{organization={Eastern Switzerland University of Applied Sciences},city={St. Gallen}, country={Switzerland}}

%% Abstract
\begin{abstract}
\input{01_abstract}
\end{abstract}

% Placeholder; for the submission, the entries 'graphicalabstract' and 'highlights' must be uncommented and filled out

\begin{highlights}
\item 
Treatment Recommender trained and validated on an RCT
%\item Clinical Decision Support
%We propose a treatment outcome predictor for the recommendation of treatments that have not yet been deployed clinically, but have already been validated in a Randomized Clinical Trial (RCT).
%Since the assignment of patients to treatments in an RCT is random, while treatment recommendation in a clinical decision support is not random, we introduce a mechanism that compensates for the missing rationale in patient assignment on the ground truth data.
\item 
Validating a treatment recommender despite missing verification evidence
\item
Counterfactual treatment verification
%Since the only data available for model evaluation are the RCT data, we face the challenge of 'missing verfication evidence': for any patient, we only know the outcome for the treatment assigned to him/her, and this assignment was random. To deal with this challenge, we introduce the concept of 'counterfactual treatment verification' that forms the basis for our evaluation procedure.
%\item Since the only treatment assignments available for validation of the treatment recommender are the RCT data, we perform a post hoc alignment between the treatments recommended by our new approach and the (random) treatment assignments, and validate on the improvement achieved by aligned vs misaligned recommendations.
\end{highlights}

%% Keywords
\begin{keyword}
treatment recommender \sep treatment recommendation validation \sep missing verification evidence \sep clinical decision support 



\end{keyword}

\end{frontmatter}

%% file: 01_abstract.tex
\textbf{Background:} Current clinical decision support systems (DSS) are trained and validated on observational data from the clinic in which the DSS is going to be applied. This is problematic for treatments that have already been validated in a randomized clinical trial (RCT), but have not yet been introduced in any clinic. In this work, we report on a method for training and validating the DSS core before introduction to a clinic, using the RCT data themselves. The key challenges we address are of missingness, foremostly: missing rationale when assigning a treatment to a patient (the assignment is at random), and missing verification evidence, since the effectiveness of a treatment for a patient can only be verified (ground truth) if the treatment was indeed assigned to the patient - but then the assignment was at random.

\textbf{Materials:} We use the data of a multi-armed clinical trial that investigated the effectiveness of single treatments and combinational treatments for 240+ tinnitus patients recruited and treated in 5 clinical centers.

\textbf{Methods:} To deal with the 'missing rationale' challenge, we re-model the target variable that measures the outcome of interest, in order to suppress the effect of the individual treatment, which was at random, and control on the effect of treatment in general. To deal with missing features for many patients and with a small number of patients per RCT arm, we use a learning core that is robust to missing features, and we build ensembles that parsimoniously exploit the small patient numbers we have for learning. To deal with the 'missing verification evidence' challenge, we introduce \emph{counterfactual treatment verification}, a verification scheme that juxtaposes the effectiveness of the DSS recommendations to the effectiveness of the RCT assignments in the cases of agreement, respectively disagreement, between DSS and RCT.

\textbf{Results and limitations:} We demonstrate that our approach leverages the RCT data for learning and verification, by showing that the DSS suggests treatments that improve the outcome. The results are limited through the small number of patients per treatment; while our ensemble is designed to mitigate this effect, the predictive performance of the methods is affected by the smallness of the data.

\textbf{Outlook:} We provide a basis for the establishment of decision supporting routines on treatments that have been tested in RCTs but have not yet been deployed clinically. Practitioners can use our approach to train and validate a DSS on new treatments by simply using the RCT data available to them. More work is needed to strengthen the robustness of the predictors. Since there are no further data available to this purpose, but those already used, the potential of synthetic data generation seems an appropriate alternative.

%% file: 1_introduction.tex
There is a surge on AI-based clinical decision support systems (DSS), and much research on their potential but also on the ways of validating them \citep{ChenEtAl:sysrev:AIIM2023,%
	%CombiEtAl:XAImanifesto:AIIM2022,
	LeiserEtAl:review:AIIM2023}.
In their meta-analysis on the evaluation of improvements achieved by clinical decision systems, \cite{KwanEtAl:BMJ2020} concentrated on studies that use controlled clinical trials. However, they 'defined a clinical decision support system as any on-screen tool designed to improve adherence of physicians to a recommended process of care' and accordingly ' focused on improvements in processes of care (eg, prescribing drugs, immunisations, test ordering, documentation), rather than clinical outcomes.' This definition restricts DSS to already deployed tools. However, as medical research progresses, it seems reasonable to include innovative treatments among the suggestions of a DSS, as soon as these treatments have been found to be beneficial. The typical way of assessing the effects of treatments is a Randomized Clinical Trial (RCT), but this implies that the only data available to train and validate a model that recommends such treatments are the RCT data themselves.

In this study, we propose a method that predicts the expected patient improvement for each of alternative treatments, where the treatment data come from a multi-armed Randomized Clinical Trial (RCT). For treatments that have been studied in an RCT and are now mature for deployment, there are no observational data to train the DSS models on, and, obviously, no independently sampled data for their validation. This leads to the challenge of training, testing and validating the machine learning core of a DSS, when the only available data to this purpose are those of the RCT itself. This leads to following challenges:
% Albeit RCT data are the gold-standard for the comparison of treatments and the identification of their effects on patients, they are less ideal when it comes to training, testing and validating a DSS that ranks these very treatments. In particular, following challenges must be addressed.

%\subsection{Framing the DSS learning$+$validation problem}
%We face three challenges of missingness.

\paragraph{A. Missing rationale for treatment assignment} In typical clinical data, the treatment that a patient receives can be relied upon as the best choice that the clinician could make, given the patient data at baseline. In an RCT, however, the assignment of patients to treatments is random. Therefore, the treatment outcome predictor should learn only the degree to which a patient improves given a treatment and in comparison to other patients, and not learn an association between patient characteristics and treatment assignment.
%We address this issue by introducing the concept of \textit{treatment appropriateness} and conducting the training and validation worklfow accordingly, cf. section \ref{subsec:mrationale}.

\paragraph{B. Missing verification evidence} Whenever we predict the expected outcome for a treatment the patient did not receive during the RCT, there are no data to quantify the quality of the prediction.
% for a patient a different treatment than the one the patient received as part of the randomized treatment assignment procedure, then there are no data to validate the recommendation. 
%We addressed this issue by introducing the concept of \textit{counterfactual treatment} and conducting the training and validation workflow accordingly, as described in section \ref{subsec:counterfactualT}.

\paragraph{C. Missing evidence} While it is necessary to train the best performing models, the analysis of historical data reveals that different clinics differ in the exact features captured regarding a patient, the exact questionnaires that are collected, the order they are collected in, and that the likelihoods of missingness at the feature and the questionnaire level. Therefore, we need to deliver predictions even when some features and even whole questionnaires are missing.
%The design decisions emanating from the need to learn with missing features (item i) are presented in section \ref{subsec:mfeatures}.

Next to missing features, we need to account for the fact that the available data are unevenly distributed among the treatments. This fact reflects reality, where some treatments are more likely to be offered than others. Even in an RCT, it is possible that not all participants are eligible for all RCT arms. In the RCT we used for our experiments, some treatments involved the usage of hearing aids, so that the assignment of patients to arms was stratified accordingly. When coming to model validation, this fact implies that there are not enough data for both training and testing.

\paragraph{Research Questions} These three challenges translate to following research questions for training, testing and validating a DSS that ranks treatments on the expected improvement they will lead to if given to a patient:
\begin{enumerate}
    \item How to deal with missing evidence when training a DSS core or RCT data?
    \item How to deal with the missing rationale in treatment assignment when training and testing the DSS on RCT data?
    \item How to deal with missing verification evidence when testing the DSS on RCT data?
    \item How to validate a DSS against an RCT dataset?
\end{enumerate}

To address these research questions we propose \kmddss{}, a method that predicts for each treatment the improvement in a patient's condition that can be expected if the patient receives that treatment, and then ranks the treatments on amount of expected improvement. The treatments come from a multi-armed RCT, so we simulate the scenario of decision support for the deployment of treatments that have been studied in an RCT only. Our method encompasses a re-design of the learning task to deal with the missing rationale and with inadequate evidence for some treatments, a counterfactual treatment verification mechanism to deal with the missing verification evidence problem, and a validation procedure based on alignment between the treatment assignments of the RCT and the top-1 treatment recommended by our model for each patient.

The rest of the paper is organized as follows. In section \ref{sec:relwork}, we discuss the role of AI for clinical decision support, with emphasis on the validation of models. In section \ref{sec:materials}, we outline the design of the Randomized Clinical Trial (RCT), the data of which we use in the examples and for our experimental evaluation. Section \ref{sec:ourmethod} describes our approach and section \ref{sec:validation} the validation procedure. We present our results in Section \ref{sec:experiments} and close with a Discussion and Outlook section \ref{sec:discussion}.

%% file: 2_relwork.tex
We first outline the contexts in which AI for clinical decision support is being studied. We then concentrate on the validation of AI-induced models, we distinguish between internal and external validation, and we focus on the challenges of the latter.
 
%\begin{itemize}
%    \item Clinical DSS design and requirements
%    \item Methods for the validation of a clinical DSS \hfill [top-1 priority]
%    \item Clinical decision support for multi-component treatments
%\end{itemize}
%
%\todoin[color=yellow]{
%This is a test citation in different forms, namely 'cite' \cite{SchleicherEtAl:ESWA2024} vs 'citet' \citet{SchleicherEtAl:ESWA2024} vs 'citep' \citep{SchleicherEtAl:ESWA2024}. It turms that 'cite' is set to 'citet' and is appropriate for saying 'authors (2024) have proven that' while 'citep' should be used for saying 'In (authors, 2024) it was proven that'.
%}

\paragraph{AI for clinical decision support}
%In their scoping review on 'ethical, legal, and social considerations of AI-based medical decision support tools', \cite{CartolovniEtAl:IJMI2022} found that 'The most prevalent issues are patient safety, algorithmic transparency, lack of proper regulation, liability \& accountability, impact on patient-physician relationship and governance of AI empowered healthcare.'

%The role of AI for (clinical) decision support
The AIIM journal itself counted almost 120 articles on this subject in year 2023, whereby the publications can be roughly categorized into: (i) articles that elaborate on the potential of a specific AI solution for decision support with respect to a specific condition, see e.g. the investigation of \cite{ItzhakEtAl:AIIM2023} on predicting episodes of hypertension, or the more recent works of \cite{JafarEtAl:AIIM2024} on the administration of diabetes and of \cite{ChenEtAl:AIIM2024} on sequenced dental plans; (ii) reviews on insights won when using AI for prediction or monitoring of a condition, such as the meta-analysis of \cite{XingEtAl:metareview:AIIM2023} on predicting response to radiotherapy by patients with lung cancer or the systematic review of \cite{SumnerEtAl:sysrev:AIIM2023} on the role of AI in physical rehabilitation; (iii) articles on the potential of a specific type of AI solutions in exploiting new types of data, e.g. the work of \cite{WassermanEtAl:AIIM2023} on causally validating treatment effects from registry data, or the approach of \cite{MichalowskiEtAl:AIIM2023} on knowledge graphs for the revision of clinical practice guidelines; and (iv) an increasing amount of contributions on physician-AI interaction, including process mining solutions \citep{LeemansEtAl:AIIM2023} and reviews thereof \citep{ChenEtAl:sysrev:AIIM2023}, human-AI collaboration protocols \citep{CabitzaEtAl:AIIM2023} and the large field of explainable AI for medicine, for which a manifesto was proposed by \cite{CombiEtAl:XAImanifesto:AIIM2022}.

The validation of AI-based decision support solutions emerges directly or indirectly on all these works. While independently designed randomized clinical trials (RCTs) are considered the most appropriate form of validation, there is substantial literature on further validation mechanisms, not least because of the costs (and duration) of RCTs, as discussed hereafter.

\paragraph{Validation of AI-based decision support}
In their systematic review and meta-analysis, \cite{KwanEtAl:BMJ2020} report on their findings from 108 studies which investigated the effects of clinical decision support systems. Of these studies, 94 were randomized. \cite{BicaEtAl:2021} elaborate on ways of using observational data to assess individualized treatment effects. \cite{WassermanEtAl:AIIM2023} investigate ways of validating on observational data in a more general context, while \cite{FranklinEtAl:2020} and \cite{WheatonEtAl:2023} highlight the potential of (nonrandomized) real-world evidence in supporting regulatory decision making.
% and propose workflow for the replication of randomized clinical trials.

When moving from clinical decision support in general towards AI-driven DSS, challenges proliferate. \cite{SuttonEtAl:NPJ2020} propose ways of alleviating 'potential harms' like alert fatigue, user distrust and dependency on computer literacy, and stress the need for a 'System for measurement and analysis of CDSS performance.' \cite{VaseyEtAl:2022} propose a 'reporting guideline for the early-stage clinical evaluation of decision support systems driven by artificial intelligence':'early \emph{live} clinical evaluation' constitutes a core piece of this guideline, whereby prospective cohort studies and non-randomized controlled trials are suggested as possible implementations. In that context, the evaluation settings and the clinical workflow/care pathway in which the AI system was evaluated must be described as part of the implementation \citep[Table 2]{VaseyEtAl:2022}.

%In their work on 'Knowledge acquisition, synthesis, and validation: a model for decision support systems', \cite{ONeillEtAl:2004} report on the development of a \emph{prospective} decision support system prototype and point out that this prototype 'is being constructed on rules and cases generated by the best available evidence.' 
Such validation procedures are needed \emph{after} the AI-based DSS is deployed. An earlier validation is needed though, namely after the AI-core is built and \emph{before} it becomes part of clinical practice. 
%\cite{MathewsEtAl:NPJ2019} stress the importance of an 'independent evaluator'.

\paragraph{Internal validation and validation in interaction with experts}
As pointed out in the systematic review of \cite{TavazziEtAl:sysrev:AIIM2023} 'the overwhelming majority of included studies [resort] to internal validation only'.\citep{TavazziEtAl:sysrev:AIIM2023} refer to articles on a specific medical condition, but the usage of existing data for validation is widespread, see e.g. \cite{XingEtAl:metareview:AIIM2023} for another medical condition. XAI-oriented approaches like \citep{CabitzaEtAl:AIIM2023,ChariEtAl:AIIM2023,MichalowskiEtAl:AIIM2023} investigate how the induced models are perceived by human experts or agree with expert knowledge or clinical guidelines. Both the XAI methods and the methods evaluating on observational data assume that the treatments under study are already practised clinically. For new treatments that have been 'only' studied inside an RCT, such a validation is not feasible.

\paragraph{External validation}
Validation without observational data demands dedicated protocols or simulators.
\cite{ChenEtAl:2023} propose a workflow for the validation of three clusters of unipoloar depression symptoms (created by a machine learning algorithm) on existing clinical trial data. \cite{GrolleauEtAl:medrxiv2023} use data from two randomized controlled trials for the validation of models built on MIMIC III data. %\cite{McLernonEtAl:2023} elaborate on the validation procedure for survival models on existing data -- training on one dataset and validating on another dataset with the same properties

\cite{JafarEtAl:AIIM2024} evaluate against the state of the art algorithms with data from a 'validated simulator' that generates suboptimal values for comparison. Simulations are also used by \cite{LeemansEtAl:AIIM2023} -- in the form of simulated processes. \cite{WhittonHunter:AIIM2023} develop an own gold standard to evaluate the proposed method. \cite{LiuEtAl:AIIM2024} evaluate on observational data (from MIMIC-IV) but also use an external validation procedure based on 'action similarity rate and relative gain'; our original validation protocol originates also from the idea of validating on relative gain.
%, but we validate against an RCT rather than against a mechanism that chooses the optimal treatment for each patient.

All aforementioned studies concerning validation of AI models intended for decision support share a common assumption: there are independently collected data that can be used for validation. This assumption does not hold for newly introduced treatments, which are themselves investigated in the context of an RCT. For our treatment recommendation method, the treatments come from an RCT and the only existing data to train, test and validate the method are the data of this RCT.

%% file: 3_materials.tex
The data used for the evaluation of our \kmddss{} come from a 10-arm Randomized Clinical Trial described in \citep{SchoisswohlEtAl:TRIALS2021,UNITIrct:Trials2023,SchoisswohlEtAl:medrxiv2024}. Hereafter, we outline the RCT, list the features we used and introduce some terminology.

\subsection{The RCT and its 'primary outcome measure'}
\label{subsec:RCT_target}
Quoting from the abstract: 'Within the Unification of Treatments and Interventions for Tinnitus Patients project, a multicenter, randomized clinical trial is conducted with the aim to compare the effectiveness of single treatments and combined treatments on tinnitus distress (UNITI-RCT). Five different tinnitus centers across Europe aim to treat chronic tinnitus patients with either cognitive behavioral therapy, sound therapy, structured counseling, or hearing aids alone, or with a combination of two of these treatments, resulting in four treatment arms with single treatment and six treatment arms with combinational treatment.' Hereafter, we use the same acronyms for the treatments as in \citep{SchoisswohlEtAl:TRIALS2021}, namely: CBT for 'cognitive behavioral therapy', HA for 'hearing aids', SC for 'structured counseling', and ST for 'sound therapy'.

The \emph{primary outcome measure}, i.e. the variable used to measure the response of patients to treatment, is the score of the Tinnitus Handicap Inventory 'THI', cf. \citep{THI:1996}, \citep{thi}. For the THI score, smaller values are better, hence a positive value of the difference $\Delta$THI between the THI score at baseline and the THI score in the final visit (i.e. immediately after treatment end) indicates that the patient improved after treatment. It is stressed that all treatments (single and combinational ones) were designed to have the same duration in total, hence the timepoint at which the THI was measured at the final visit was independent of the treatment / RCT arm.

%is considered as a positive response. For our analyses, we consider prediction of the THI score at the final visit, and of the $\Delta$THI.
%It is noted that a THI score of at least 18 at screening, i.e. at the beginning of recruitment, was set as inclusion criterion for participation to this RCT.

\subsection{RCT participants}
\label{subsec:RCTparticipants}
Each of the 5 clinical centres involved in the RCT planned to recruit 100 patients, summing up to 500 patients total \citep{SchoisswohlEtAl:TRIALS2021}. The RCT was closed in December 2022.
%with a total of 461 participants. 
%\todo[color=moreBGcolor]{For WINNY: why 422 and not 461?}
For our analyses, we used the data of 376 (out of 461) participants that had a THI score in the final visit. There is a slightly uneven distribution among the RCT arms: main reasons were differences in the dropout rates, as well as the fact that the arms involving hearing aids (HA) could be used only for participants with hearing loss above a given threshold \footnote{The process of assigning participants with hearing loss to arms is described under 'Randomization and blinding' in \citep{UNITIrct:Trials2023} Details on the actual data are in \cite{SchoisswohlEtAl:medrxiv2024}}. From these participants, 75\% were used for training and 25\% were held out for testing.

\subsection{The RCT features used in our analyses}
\label{subsec:RCTfeatures}
\begin{figure}[htb]
\label{fig:patientRepresentationInReducedFS}
\includegraphics[scale=0.5]{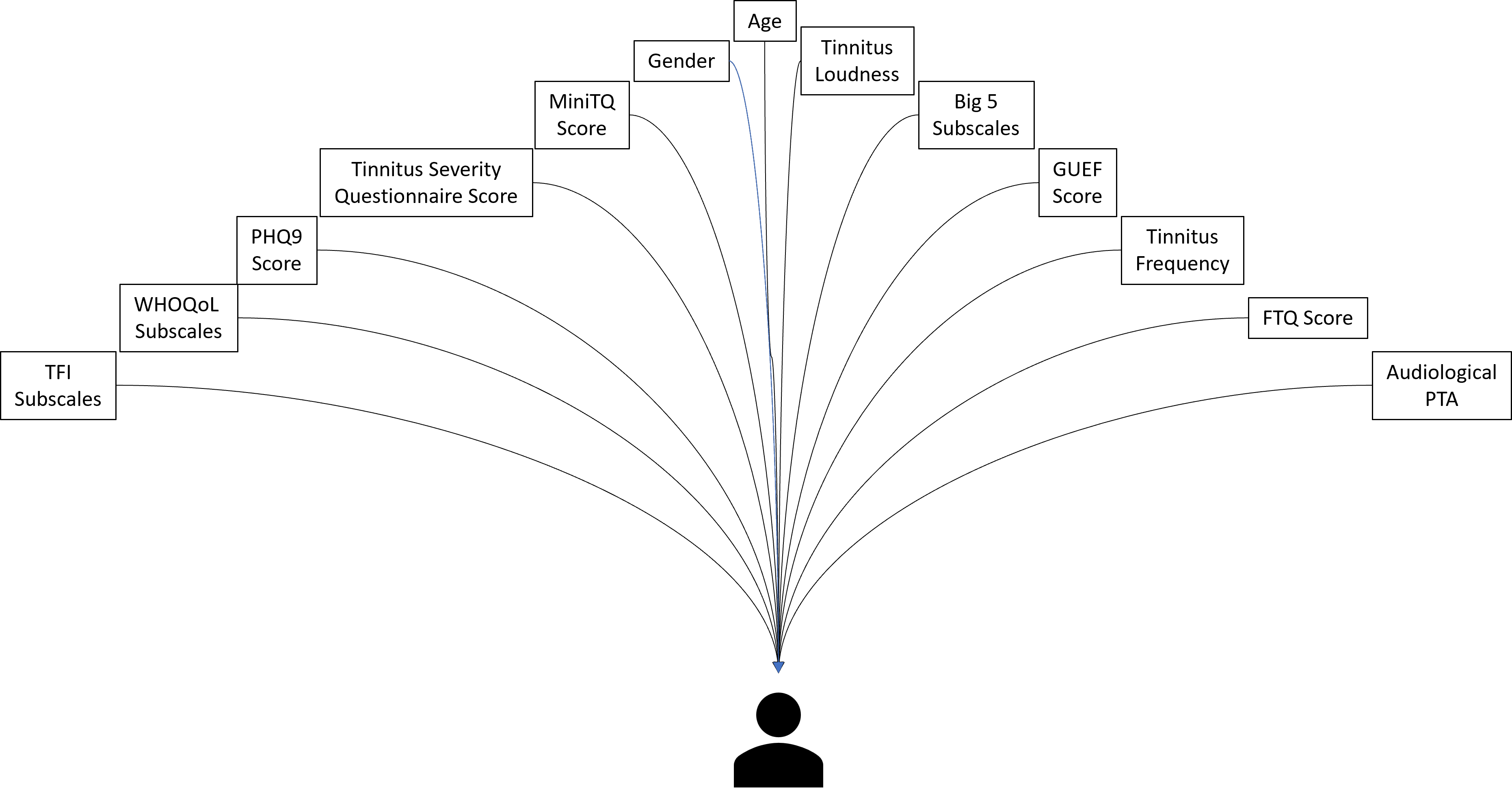}
\caption{Patient representation in a high-dimensional feature space encompassing single features, subscales and questionnaire scores}
\end{figure}

The RCT dataset encompasses a comprehensive array of questionnaires and audiological measurements. The full list of assessments (recorded at baseline, interim visit, final visit and follow up) appears as Table 4 in \citep{UNITIrct:Trials2023}. In clinical practice though, it is possible and even likely that a patient will not fill all the questionnaires. Hence, the features considered in \kmddss{} are the following features as recorded at baseline, whereby any of them may be missing:
%\todo[color=MyraBGcolor]{Add citations}
\begin{itemize}
\item \textit{Single features:} gender, age, tinnitus loudness, tinnitus frequency, audiogram
\item \textit{Scores of} questionnaires, namely the 'Fear of Tinnitus Questionnaire' (FTQ) cf. \citep{ftq}, the 'Mini Tinnitus Questionnaire' (Mini-TQ) \citep{minitq}, the 'Questionnaire on Hypersensitivity to Sound' (G\"UF) cf. \citep{HypersensitivityToSound}, the 'Patient Health Questionnaire for Depression' (PHQ-D / PHQ-9) cf. \citep{phq9}, and the 'Tinnitus Severity Questionnaire' (TSQ) cf. \citep{tinnitusSeverity}
\item \textit{Scores of the subscales within the questionnaires:} 8 subscales of the 'Tinnitus Functional Index' (TFI) cf. \citep{tfi}, 4 subscales of the the 'World Health Organization Quality of Life abbreviated' (WHOQoL-Bref) cf. \citep{whoqlbref}, and the 5 subscales of the 'Big Five Inventory 2' (BFI-2) cf. \citep{BFI-2}
\end{itemize}
summing up to 32 features, which are summarized in Figure \ref{fig:patientRepresentationInReducedFS}. It is noted that the THI Score at baseline was omitted because it would overpower the other features.
% \todoin[color=anyBGcolor]{
% \begin{list}{$\circ$}{}
% \item Number of features total?
% \item Number of subscales for TFI, WHOQoL-Breaf and BFI-2 ?
% \item In Figure \ref{fig:patientRepresentationInReducedFS}, replace TSQ with 'TSQ score'
% \end{list}
% }

\paragraph{Notational convention}
Since CBT, HA, SC and ST are used as treatment delivered alone or as components of a combinational treatment, we use the word \textbf{'therapy'} when referring to CBT offered alone or as one of two treatment components, i.e. for all arms that involved CBT -- and similarly for HA, SC and ST.

%% file: 4_ourDSScore.tex
\begin{figure}[htb]
\label{fig:bigpic}
\centerline{\includegraphics[page=1,scale=0.5]{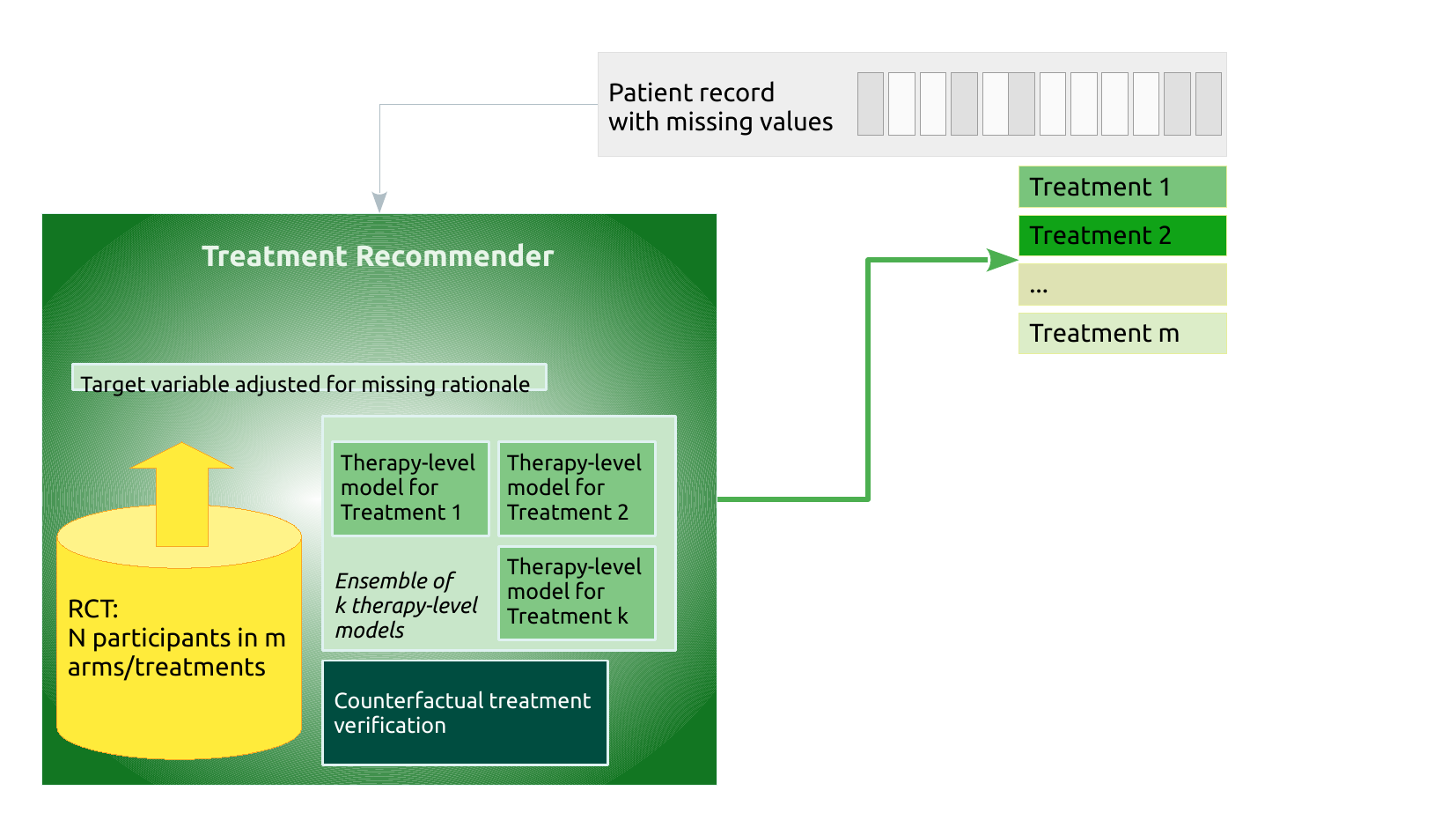}}
\caption{\kmddss{} for the ranking of treatments on improvement after adjusting the target variable and performing counterfactual treatment verfification: the left part of the figure refers to model learning on the RCT data, whereupon predictions per treatment are performed by an ensemble of therapy-level models; the right part delivers a recommendation for each patient, comprised of the treatments ranked on expected improvement (darker green colors are better); in the patient's record, white boxes refer to missing features
%; the validation procedure aligns for each patient the top-1 ranked treatment of the \kmddss{} to the treatment this patient received during the RCT (right lowermost part of the figure)
}
\end{figure}

To support a clinician' choice of treatment for a patient $x$ from a list of treatment options $\mathcal{T}=\{T_1,\ldots,T_m\}$, we propose \kmddss{}, an algorithm that for each $T\in\mathcal{T}$ predicts the expected outcome $\widehat{O(x,T)}$ of $T$ for patient $x$ and then ranks the treatments in $\mathcal{T}$ on expected outcome, while also delivering its own confidence on each prediction. Our approach is outlined in Figure \ref{fig:bigpic} and described hereafter, paying particular emphasis on how we deal with the forms of missingness that were listed in the first three research questions.
%Our approach is depicted on Figure \ref{fig:bigpic} and the methods it uses to address the missingness challenges are described thereafter.

\subsection{Modeling the target variable
%expected treatment outcome 
under the 'missing rationale' premise}
The expected \emph{outcome} after a given treatment $T$ can be modeled as (i) a binary \emph{response variable} (Y/N) that expresses whether the patient responded to treatment or not, as (ii) a quantity that measures the patient's condition after treatment or as (iii) improvement with respect to that quantity, comparing the quantity's value at baseline, i.e. before treatment, and the value after treatment $T$ was completed.

To facilitate the recognition of treatments that do not lead to improvement (and therefore should be omitted), we choose as target variable the option (iii), i.e. the difference between quantities, and thus compute for each patient $x$ and treatment $T\in\mathcal{T}$ the \emph{expected treatment outcome} as:
\begin{equation}
\label{eq:deltaQ}
\Delta{}Q(x,T)= 
Q(x,\_,t_{\mathit{baseline}}) - \widehat{Q}(x,T,t_{\mathit{treatmentEnd}})
\end{equation}
where $Q()$ is the quantity measuring the patient's condition, while $t_{\mathit{baseline}}$ and $t_{\mathit{treatmentEnd}}$ denote the two time points where $Q()$ is measured, namely before treatment start and at treatment end.

It is stressed that the value of $Q()$ at baseline is independent of the treatment (as denoted by the \_ at the second position), while the value of $Q()$ at treatment end is estimated by \kmddss{} (as denoted by the $\widehat{Q}(\ldots)$). In the following, we simplify the notation by skipping the argument $t_{\mathit{treatmentEnd}}$, so that the true value after treatment is denoted as $Q(x,T)$ and the estimated one as $\widehat{Q}(x,T)$.

Since the assignment of a patient to a treatment / RCT arm is done randomly, the learning core of the \kmddss{} should not seek a rational link between the treatment assigned to a patient and the improvement this patient experienced. To suppress this link to the extent possible, we \textit{adjust} the $\Delta{}Q(x,T)$ of Eq. \ref{eq:deltaQ} by subtracting from it the average improvement observed among the patients in the training set who actually received treatment $T$:
%\todo[color=anyBGcolor]{For Vishnu: is the correction done during training or during ranking after the prediction?}
\begin{equation}
\label{eq:adjustedDeltaQ}
\Delta_{adj}Q(x,T) = \Delta{}Q(x,T) - avg_{D_{train},T}\{\Delta{}Q\}
\end{equation}
where the average is computed over the subset of patients in the training set, who received treatment $T$: 
\begin{equation}
\label{eq:avgImprovement}
avg_{D_{train},T}\{\Delta{}Q\} =
\frac{\sum_{(u,T)\in{}D_{train}}
\left(
Q(u,\_,t_{\mathit{baseline}}) - Q(u,T)
\right)
}{|\{(u,T)\in{}D_{train}\}|}
\end{equation}

% \begin{equation}
% \label{eq:adjustedDeltaQ}
% \Delta_{adj}Q(x,T) = \Delta{}Q(x,T) - avg_{D_{T}}\{\Delta{}Q\}
% \end{equation}
% where the average is computed over the patients who received the same treatment $D_{T}$ as:
% \begin{equation}
% \label{eq:avgImprovement}
% avg_{D_{T}}\{\Delta{}Q\} =
% \frac{\sum_{(u,tr=T)\in{}D_{train}}
% \left(
% Q(u,\_,t_{\mathit{baseline}}) - Q(u,T)
% \right)
% }{|D_{tr=T}|}
% \end{equation}

Note that in Eq. \ref{eq:avgImprovement}, the value $Q(u,tr)$ is the true value, since it is computed for patients in the training set $D_{train}$. Accordingly, the $avg_{D_{train}}\{\Delta{}Q\}$ is also a ground truth value. In contrast, the adjusted expected treatment outcome \emph{is} an estimate, cf. Eq. \ref{eq:deltaQ}.

%\todoin[color=yellow]{TO REMOVE: \newline
%To deal with the fact that the treatments have been assigned to the patients randomly, \kmddss{} learns patient improvement in comparison to other patients and thus predicts treatment \textit{appropriateness}: instead of predicting for a patient and a treatment the expected treatment outcome improvement, we predict  
%\(
%\Delta{}THI - average(\Delta{}THI)
%\)
%where the second term is the average difference in the THI score among all patients who received that treatment during the RCT. We term this adjusted target variable ‘treatment appropriateness’: a treatment is deemed appropriate if a patient’s THI score improves more than the average improvement achieved through random assignment of this treatment.
%}

%In our experiments, we use as $Q()$ the primary outcome measure of the RCT, cf. section \ref{subsec:target}. Without loss of generality, we assume that smaller values of $Q()$ are better (as is the case for the primary outcome variable THI of the RCT in section \ref{sec:materials}), so that positive values of $\Delta{}Q(x,T))$ indicate improvement. Accordingly, \kmddss{} ranks the values $\Delta{}Q(x,T_i), i=1\ldots{}k$ in descending order. 

\subsection{Learning under the 'missing verification evidence' premise: counterfactual treatment verification}
%\subsection{Evaluation despite the missing verification evidence: Counterfactual treatment evaluation method}
\label{subsec:counterfactualT}
As depicted in Figure \ref{fig:bigpic}, \kmddss{} receives as input the data of a patient $x$ predicts the (adjusted) expected treatment outcome for each treatment $T\in\mathcal{T}$. To train and test a recommender on these predictions, we 
need the true treatment outcome for each treatment. However, within the RCT, each patient received exactly one treatment, so that the verification evidence for all other treatments is missing. To deal with this issue, we introduce the concepts of \emph{counterfactual score} and \emph{counterfactual treatment verification} as follows:
% \todo[color=anyBGcolor]{For Vishnu: \textit{Verification} instead of \textit{Evaluation}}

%\todo[color=MyraBGcolor]{Use the term 'Nearest Counterfactual Neighbor's Improvement' in the big picture}
%
\paragraph{Counterfactual score for each treatment}
Given is a patient $x$ who received treatment $T_x$. For each other treatment $T\in\mathcal{T}\setminus\{T_x\}$ we compute the \emph{counterfactual score} from the data of the patients who actually received treatment $T$ and are most similar to patient $P$, as described in the pseudocode of Table \ref{tab:counterfactualscore}. In this table, $s()$ denotes the similarity function used to identify the nearest neighbours of $P$.

\begin{table}[htb]
\small
\begin{tabular}{cp{0.85\textwidth}}
\hline
& \textbf{Counterfactual score} for patient $x$ and treatment $T\neq{}T_x$ \\
\hline
1 & Let $\mathit{arm}(T)$ be the set of patients who received treatment $T$
\\
2 & For each $u\in\mathit{arm}(T)$: compute the similarity of $u$ to $x$, $s(u,x)$ 
\\
3 & Compute the set $NN(k,x,\mathit{arm}(T))$ of the $k$ nearest neighbours to $x$ among the patients in $\mathit{arm}(T)$
\\
4 & Derive the \emph{counterfactual score} of $Q$ for $T$ on patient $x$ as:
\begin{equation}
\label{eq:counterfQ}
\tilde{Q}(x,T) =
\frac{1}{k}\sum_{u\in{}NN(k,x,\mathit{arm}(T))}s(u,x)\cdot{}Q(x,T)
\end{equation}
\\
\hline
\end{tabular}
\normalsize
\caption{\protect{\label{tab:counterfactualscore}}Pseudocode for the computation of the counterfactual score $\tilde{Q}(x,T)$ for a patient $x$ and each treatment that this patient did \emph{not} receive}
\end{table}

\paragraph{Matrix completion with counterfactual verification evidence} Let $M$ be the matrix of true treatment outcome values for the set of $N$ patients who participated in the RCT and for all treatments / RCT arms. $M$ has as many columns as treatments and for each patient, only one of the columns has a non-null value -- the column corresponding to the RCT arm, to which the patient was assigned. We use the counterfactual scores to fill the rest of the matrix, as shown on Figure \ref{fig:fillingthematrix}.

\begin{figure}[htb]
\begin{minipage}{0.45\textwidth}
\tiny
\begin{tabular}{ccccc}
\hline
%\textbf{Patient} 
& \multicolumn{4}{c}{\textbf{Treatment/RCT arm}} \\
\textbf{ID} & $T_1$ & $T_2$ & \ldots & $T_{k}$ \\ 
\hline
\#1 & $Q(\#1,T_1)$ & -- & \ldots & -- \\
\#2 & -- & -- & \ldots & $Q(\#2,T_{k})$ \\
\#3 & $Q(\#3,T_1)$ & -- & \ldots & -- \\
\ldots & & & & \\
\#N & -- & $Q(\#N,T_2)$ & \ldots & -- \\
\hline
\end{tabular}
\normalsize
\end{minipage}
\hfil$\rightarrow$\hfil
\begin{minipage}{0.45\textwidth}
\tiny
\begin{tabular}{ccccc}
\hline
%\textbf{Patient} 
& \multicolumn{4}{c}{\textbf{Treatment/RCT arm}} \\
\textbf{ID} & $T_1$ & $T_2$ & \ldots & $T_{k}$ \\ 
\hline
\#1 & $Q(\#1,T_1)$ & $\tilde{Q}(\#1,T_2)$ & \ldots & $\tilde{Q}(\#1,T_k)$ \\
\#2 & $\tilde{Q}(\#2,T_1)$ & $\tilde{Q}(\#2,T_2)$ & \ldots & $Q(\#2,T_{k})$ \\
\#3 & $Q(\#3,T_1)$ & $\tilde{Q}(\#3,T_2)$ & \ldots & $\tilde{Q}(\#3,T_k)$ \\
\ldots & & & & \\
\#N & $\tilde{Q}(\#N,T_1)$ & $Q(\#N,T_2)$ & \ldots & $\tilde{Q}(\#N,T_k)$ \\
\hline
\end{tabular}
\normalsize
\end{minipage}
\caption{\protect{\label{fig:fillingthematrix}}Filling the matrix of treatment outcome values for all treatments and all patients: the original matrix contains only one filled value per patient (left subfigure); the filled matrix contains $k-1$ counterfactual scores per patient (right subfigure)}
\end{figure}

%\paragraph{Example:} The procedure is depicted on Figure 5 for the example of a patient who was assigned to the CBT arm during the RCT and whose counterfactual treatment outcome must be computed for the treatment HA.
%
%\begin{figure}[htb]
%\label{fig:counterf}
%\caption{Computing the outcome of a counterfactual treatment from similar patients who received that treatment: for the given patient (uppermost left corner), the workflow includes (i) identification of similar patients who received the treatment under consideration, here HA, and (ii) computation of the average adjusted DTHI over these patients,while weighting it on similarity to the patient under study,  (lowermost right corner of the figure)}
%\end{figure}

%\kmddss{} method for counterfactual treatment evaluation first identifies the three most similar patients who had actually received treatment HA. It then takes the treatment outcome for each of these three patients, weights it with the similarity to the patient under study, and computes the average of these weighted values. 

\color{black}

\subsection{Training an Ensemble of therapy-level models under the `missing evidence' premise}
\label{subsec:therapyLevelmodels}
The prediction of outcome values for each patient and treatment can be perceived as a vector-filling task for the vector of $m$ treatments. The model is trained on the matrix at the righthand side of Figure \ref{fig:fillingthematrix} after the computation of the counterfactual treatment scores for the patients in the training set. For model learning and for model application we face two challenges. First, as depicted on Figure \ref{fig:bigpic}, some features may be missing for a patient. Next, the evidence for some treatments may be much smaller than for other treatments, e.g. because of differences in eligibility. In our example RCT described in section \ref{subsec:RCT_target}, some treatments involve hearing aids, and the likelihood that an RCT participant is eligible for them is less than 50\%.

\subsubsection{Learning with missing features}
\label{subsub:mfeatures}
To deal with the missing features challenge, we use XGBoost as learning core, a tree-based algorithm for numerical targets. XGBoost \citep{ChenGuestrin:KDD2016} is designed to segment data into groups with similar deviations from the group average in a hierarchical manner, and to generate multiple trees, whereby it performs feature subsampling for each tree \footnote{\cite{ChenGuestrin:KDD2016} use the term 'column subsampling' instead, cf. their section 2.3, but they consider a column as synonym for a feature.}. Through grouping, feature subsampling, and explicitly modeling missing values during purity computations, XGBoost inherently tolerates the absence of some features for some data records.

%making it versatile for various applications. It exhibits performance comparable to other tree-based methods like Random Forests.

%\kmddss{} concentrates on a core set of features that are likely to be recorded for each patient, but can still be missing for some patients; this core set is described in Section 3.2.1. The learning algorithm is also chosen to be robust to missingness, as described in Section 3.2.2.
%
%\paragraph{Core set of features - single features and aggregate scores}
%The RCT dataset encompasses a comprehensive array of questionnaires and audiological measurements, positioning it as a valuable resource for future clinical applications. In clinical practice though, it is possible and even likely (as we learned in WP3) that a patient will not fill all the questionnaires. Hence, the features considered for \kmddss{} are the following (cf. Figure \ref{fig:patientRepresentation}), whereby any of them may be missing: 
%
%\begin{figure}[htb]
%	\label{fig:patientRepresentation}
%	\caption{Patient representation in a high-dimensional feature space encompassing single features, subscales and questionnaire scores}
%\end{figure}

\subsubsection{Boosting the evidence for outcome prediction per treatment}
\label{subsub:therapyLevelModels}
Statistical analyses on RCT data have been designed to deal with small numbers of participants per arm. For a treatment recommender, the problem of RCT arms with very few participants is exacerbated, because the recommender needs data for training, data for tuning and data for testing. Even when using cross-validation, it can occur that there is only one patient with a given treatment in the test subsample.

To boost the evidence available for outcome prediction per treatment, we combine data from multiple arms into multiple \emph{therapy-level} models, which are used either separately or as an ensemble. More precisely:
\begin{enumerate}
\item An RCT arm corresponds to either a single treatment or a combinational treatment. We focus on the treatment `components' or `elements' in each arm, distinguishing between RCT arms with one component (a single treatment) and RCT arms with multiple components (combinational treatments). According to the notational convention mentioned at the end of Section \ref{sec:materials}, we use the term \emph{therapy} instead of `component'.
%, to honour the fact that each such component can be used as standalone treatment.
\item For each therapy, we combine data from all arms that involve this therapy as component, and train a \emph{therapy-level model}.
\item We build an \emph{ensemble of $l$ therapy-level models}, where $l<m$ and equal to the numbere of components that appear in the treatments. This ensemble that have a restricted weighted voting scheme:
	\begin{itemize}
	\item For each single treatment $T$, only one ensemble member is allowed to vote -- the therapy-level model for the therapy $T$.
	\item For each combinational treatment $T_i$ that consists of components $T_{i,1},\ldots,T_{i,j}$, with $j\leq{}l$, only $j$ ensemble members are allowed to vote - the therapy-level models for $T_{i,1},\ldots,T_{i,j}$
	\end{itemize}
whereby a vote is weighted with the confidence of the ensemble member to its prediction, and this confidence is implemented as the inverse of the mean prediction error of that ensemble member (i.e., the higher the error, the lower the weight that is applied to the prediction of that ensemble member)
\end{enumerate}

\begin{figure}[htb]
	\label{fig:therapiesAndArms}
	\centerline{\includegraphics[scale=0.4]{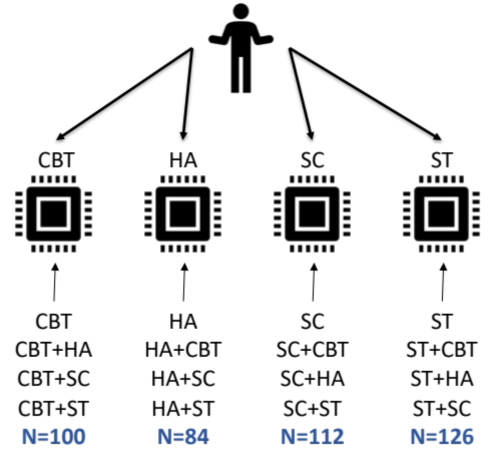}}
	\caption{Overview of the 10 RCT arms, where the treatment in an arm can is one component, e.g. CBT, or a pair of components, e.g. CBT$+$ST: as evidence for CBT we consider all arms where CBT was offered, and same for each of HA, SC and ST}
\end{figure}

\paragraph{Example} In the RCT of section \ref{sec:materials}, there are four components, namely Cognitive Behavioral Therapy (CBT), Hearing Aid (HA), Structured Counseling (SC), and Sound Therapy (ST), offered as single treatments (4 arms) and in pairs (6 arms). This results into four therapy-level models, as shown on Figure \ref{fig:therapiesAndArms}.

In this figure, the leftmost column sums up the patients that received CBT as single treatment (first row) or in combination with another component, coming to a total of 100 patients. The other three columns sum up the patients that received HA, SC and ST respectively. It is evident that an RCT arm is used more than once: the arm with components CBT and ST appears twice, in the leftmost column as 'CBT$+$ST' and in the rightmost column as 'ST$+$CBT'. It is to be noted that even though the same patient can exist in the training data of two models, the target variable as defined by each of the models will be different, as described in Equation \ref{eq:adjustedDeltaQ}.
%  \todo[color=MyraBGcolor]{Add remark in section \ref{sec:discussion} on limitations due to this overlap}

\color{MyraFGcolor}
%To address the diversity among treatments, \kmddss{} contains multiple models at therapy-level and an elaborate workflow for learning and validation, which prohibits data leakage despite the small size of the samples. Since the number of patients per arm was very small, we first estimated the effects of the primary treatments (cf. Figure 5), namely of Cognitive Behavioral Therapy (CBT), Hearing Aid (HA), Structured Counseling (SC), and Sound Therapy (ST) by inducing four models total. The sample sizes are in Figure 4.
%
%\paragraph{How are the four therapy-level models used:}
%\kmddss{} allows two ways of prediction with the four therapy-level models:
%\begin{itemize}
%\item \textit{A model alone:} it delivers predictions for the primary treatment it refers to (e.g. ST), whereby it exploits evidence from all arms where this treatment was used (cf. Figure 4)
%\item \textit{Ensemble-of-four:} it delivers predictions for each single treatment and for each pair of treatments, whereupon the vote of each of the four ensemble members is weighted by the member’s confidence to its decision
%In the present implementation of the DSS front-end, only the ensemble-of-four is used. However, if a clinic wants to concentrate only on some of the treatments, the predictions of the other treatments can be ignored.
%\end{itemize}

\color{black}

\section{Validation procedure for \kmddss{}}
\label{sec:validation}
Since the assignment of patients to an RCT arm is at random, while the \kmddss{} assigns treatments not at random, an intuitive basis for evaluation is to study the treatment improvement when the RCT assignment agrees with the \kmddss{} assignment vs when the two assignments disagree. We first describe this procedure, then explain why it is too restrictive in the context of clinical decision support, and propose an enhanced validation procedure.

\subsection{Validation by post hoc alignment between treatment recommender and RCT}
The core idea of the original validation procedure of the UNITI RCT was that a DSS recommendation for a treatment is good if the patients who actually received this treatment experienced more improvement than other patients. More formally,
%\todo[color=moreBGcolor]{FOR WINNY: Okay so?}
all patient assignments to treatments, as recommended by \kmddss{}, were marked as
\begin{itemize}
\item \emph{aligned (+):} The treatment recommended to the patient happens to be the same as the treatment the patient received in the RCT.
\item \emph{aligned (-):} The treatment recommended to the patient happens to be different from the treatment the patient received in the RCT.
\end{itemize}
where the expression 'happens to be' is intentional: the treatment a patient received was subject to randomization. This given, we expect that if the recommendation of \kmddss{} is indeed of benefit to the patient, then the treatment improvement will be higher for patients in the 'aligned (+)' group, i.e. when the assigned treatment is the recommended one, rather than for patients assigned to the 'aligned (-)' group.

Hence, to validate the recommendations of \kmddss{} we would investigate the amount of improvement within the 'aligned (+)' group vs within the 'aligned (-)' group. Additionally, it would be possible to concentrate only on patients in each group which exhibited significant improvement in the clinical sense. 

\subsection{Enhanced validation procedure}
The aforementioned procedure assumes that the DSS recommends exactly one treatment per patient. However, our \kmddss{} rather returns \emph{for each treatment} the predicted improvement together with the confidence of the model in the prediction. By doing so, a specific case of randomness is avoided, namely choosing among treatments that have very similar values of predicted improvement and condidence. Moreover and more importantly, the clinician has the opportunity to inspect the individual predicted improvement values and confidence levels, and to take account of further factors that many not be known to \kmddss{}, e.g. the availability schedule for scarce facilities that may be needed by one treatment but not by others with similar expected improvement.

The additional functionality of our \kmddss{} counteracts the validation design. To perform the validation nonetheless, we aligned the RCT arms with \kmddss{} in several steps as explained hereafter.

\paragraph{Step 1: Alignment at top-1 position}
For each patient, we ranked the treatments on the improvement predicted by \kmddss{}, and we selected the top-1, i.e. the one with the largest improvement. This corresponds to assuming that the clinician would choose the treatment with the largest predicted improvement.

\paragraph{Step 2: Distinguishing between binary and ternary alignment} A treatment may have more than one components. In some cases, it is reasonable to demand that the treatment suggested by \kmddss{} is exactly the same than the one to which it is aligned -- e.g when the treatment consists of one medication to be taken every morning and one to be taken every evening. In other cases, a component may be an important but dispensable enhancement, e.g. when a medication which may cause headache as a side effect is taken together with a medication against headache. Hence, we allow for a third case of alignment, where the treatment recommended by \kmddss{} overlaps with the treatment in the RCT arm only partially.
%In the RCT data used for our evaluation, there are singleton treatments combinations of two treatments, so we can simulate the second case multi-component treatment.
Thus we distinguish among two types of alignment: 
\begin{description}
\item [\emph{Binary alignment:}] the treatment arm of the RCT is the same as the top-1 treatment in the ranked list of \kmddss{} (alignment = TRUE) or is not (alignment = FALSE).
\item [\emph{Ternary alignment:}] the treatment arm of the RCT contains one treatment component that is also contained in the the top-1 treatment of \kmddss{} (alignment = 1) or contains no common component (alignment = 0) or is fully aligned (alignment = 2).
\end{description}

\paragraph{Step 3: Building a validation set.}
For the validation of the predictions done by \kmddss{}, we use the heldout validation set. However, given that we consider two types of alignment and we concentrate on the top-1 treatment only, the heldout dataset becomes too small for validation \footnote{For our concrete RCT dataset, the holdout dataset is a random sample of 94 patients. These patients are unevenly distributed among the RCT arms, because the dropout rates differ among treatments/arms. There is even one RCT arm with only 3 patients in the heldout dataset. Such a small dataset would prohibit generalization of the findings.}

For the alignment with the RCT, we use the complete dataset of the RCT. To ensure that no unfair advantage is allotted to \kmddss{}, we check the likelihood of alignment in the heldout dataset vs complete dataset. If \kmddss{} were overfit, then it would have predicted with high likelihood the treatment that the patient actually received and the improvement that the patient actually experienced, whereupon the alignment would be disproportionaly high (i.e. close to 100\%) among the training data. 

%% file: 5_results.tex
We trained and tested our \kmddss{} on subsamples from the RCT data described in section \ref{sec:materials}.
%
%\paragraph{Sampling for training, testing and validation.}
As per standard procedure for machine learning, the first step in creating the training data for each of the above four models was to create the holdout test set on which all the final models were to be evaluated. we held out a stratified sample of 25\% of the data for validation, and used the remaining 75\% for model learning. The stratified selection ensured that the training data did not over- or under- represent treatments that were observed less frequently than others.

Model learning on the 75\% of the data encompassed training, testing and tuning of the XGBoost hyperparameters. This was done by 5-fold cross-validation on these data, i.e. without touching the holdout dataset that we reserved for validation. This means that during hyperparameter tuning each model was effectively trained on 60\% of the data (4 folds a 15\% of the data), and are not exposed to the rest. 

\subsection{Performance as prediction error}
\label{subsec:RMSE_Results}
The target variable described under Materials is the difference $\Delta$THI between the THI score at baseline and the THI score after treatment, where positive values are better, indicating that the patient improved (i.e. had a lower THI score) after treatment. For the performance of \kmddss{} we chose the Root mean Squared Error (RMSE), i.e. we sum the square of the differences over all predictions (for the data in the holdout sample) and compute the square root. RMSE penalizes predictions with large errors. Since XGBoost generates multiple subsets of data hierarchically, the final error is the averageRMSE over these subsets.

\subsubsection{Prediction error for each therapy}
\label{subsub:RMSE_perTherapy}
Table \ref{tab:RMSE_therapyLevel_Results} depicts the RMSE when predicting the $\Delta$THI of each therapy with the therapy-level models described in section \ref{subsub:therapyLevelModels}. %\todo[color=MyraBGcolor]{replace treatment-level with therapy-level in section \ref{sec:ourmethod}}

\begin{table}[htb]
\begin{tabular}{lcccc}
\hline
\textbf{Evaluation on verified evidence} & \\
\textbf{for each Therapy:} & CBT & HA & SC & ST \\
\textbf{average RMSE} & 15.5 & 14.2 & 12.6 & 17.8 \\
\hline
\end{tabular}
\caption{\protect{\label{tab:RMSE_therapyLevel_Results}}Root Mean Squared Error (RMSE) of therapy-level models over the holdout data (25\% of the data): An error of 10 units means that a patient who improved by 30 units has a prediction between 20 and 40.}
\end{table}

Table \ref{tab:RMSE_therapyLevel_Results} depicts average RMSE before addressing the issue of missing evidence. These values serve as reference for the evaluation using counterfactual treatment verification (cf. section \ref{subsec:counterfactualT}).
% \textcolor{anyFGcolor}{Nearest Counterfactual Neighbor's Improvement} 
The results are shown in Table \ref{tab:RMSE_counterfactual_Results}.
% \todo[color=anyBGcolor]{For Vishnu: \textbf{Nearest Counterfactual Neighbor’s Improvement} replaced. New term okay? }

\begin{table}[htb]
\begin{tabular}{lcccc}
\hline
\textbf{Evaluation with} & \\
\textbf{counterfactual treatment verification} \\
\textbf{for each Therapy:} & CBT & HA & SC & ST \\
\textbf{average RMSE} & 10.4 & 12.9 & 11.5 & 12.0 \\
\hline
\end{tabular}
\caption{\protect{\label{tab:RMSE_counterfactual_Results}}Average RMSEfor the therapy-level models using the counterfactual treatment verification over the holdout test data: All errors are lower than the predictions over actual ground truth, suggesting that the models can be trusted}
\end{table}

To interpret the results in Table \ref{tab:RMSE_counterfactual_Results}, it must be stressed that a low average RMSE value does not guarantee that the model is good, but a high error would mean that the model is not to be trusted. We therefore use the average RMSE values in Table \ref{tab:RMSE_therapyLevel_Results}, i.e. the error under verified evidence, as reference to interpret the values in Table \ref{tab:RMSE_counterfactual_Results}. When juxtaposing the average RMSE value per treatment in the two tables, we see that the error when evaluating on counterfactuals is lower than the error for the verified evidence, i.e. for the treatment assigned by the RCT (at random) for all four therapy-level models. This suggests that models tested against the  counterfactual scores can be trusted, since the randomization of treatment in the RCT means that there is only a 10\% chance that the treatment assigned to a patient is the ideal treatment.

\subsubsection{Prediction error for each treatment in the RCT}
\label{subsub:RMSE_perArm}
The RMSE per treatment is depicted in the Table \ref{tab:RMSE_armLevel_Results}, whereby the prediction for each pair of treatments (TreatmentX + TreatmentY) comes from two of the four therapy-level models, hence two RMSE values are delivered. It must be stressed here that for the evaluation we consider only those patients in the test subsample that were actually assigned to this specific arm. For some treatments, the number of such patients is very low, especially
%It is possible, and actually occurred, that the number of such patients in the test subsample is very small, hence no RMSE could be computed. In general, this is more likely
for treatments that have many dropouts, as well as for treatment arms that involved hearing aids, since not all RCT participants were eligible for them.
%This implies that the average RMSE may be accompanied by a large variance.

\begin{table}[htb]
\begin{tabular}{lcccc}
\hline
\textbf{Evaluation} & \multicolumn{4}{c}{\textbf{Average RMSE of Therapy-level predictors}} \\
\textbf{for each} & CBT & HA & SC & ST \\
\textbf{arm/treatment:} & predictor & predictor & predictor & predictor \\
\hline
CBT & 11.79 &
\\
CBT$+$HA & 7.3 & 6.48
\\
CBT$+$SC & 11.19 & & 14.72
\\
CBT$+$ST & 15.51 & & & 12.24
\\
HA & & 12.86
\\
HA$+$SC & & 5.46 & 4.29 
\\
HA$+$ST & & 9.11 & & 13.03
\\
SC & & & 8.51
\\
SC$+$ST & & & 13.54 & 12.30
\\
ST & & & & 16.62
\\
\hline
\end{tabular}
\caption{\protect{\label{tab:RMSE_armLevel_Results}}Root Mean Squared Error from the ensemble-of-four therapy-level predictors: for each single treatment, only one therapy-level predictor delivers predictions, hence one RMSE value is returned; for each pair of treatments, two of the therapy-level predictors deliver predictions, hence two RMSE values are returned; the values are cut at the second position after the comma.}
\end{table}

%For the interpretation of the results in Table \ref{tab:RMSE_armLevel_Results}, following aspects must be taken into account:
%The ensemble of therapy-level predictors does not predict the best treatment but the expected improvement for each treatment.

As can be seen on Table \ref{tab:RMSE_armLevel_Results}, the therapy-level predictors vary in the average RMSE they achieve for the different arms: the HA predictor tends to exhibit lower average RMSE than the others, possibly indicating that the improvement of patients with hearing aids is a bit easier to predict; the SC predictor exhibits large differences, depending on the arm it predicts; the ST predictor tends to exhibit higher average RMSE than the others. When looking at the arms, treatments involving sound therapy (ST) seem more difficult to predict, and treatment involving hearing aids (HA) seem easier to predict.

\subsection{Results for binary alignment}
\label{subsec:binaryA_Results}
To ensure that the \kmddss{} takes no unfair advantage from the data, we checked the likelihood of alignment in the heldout dataset vs complete dataset. If the \kmddss{} were overfit, then it would have predicted with high likelihood the treatment that the patient actually received and the outcome improvement that the patient actually experienced, whereupon the alignment would be disproportionaly high (i.e. close to 100\%) among the training data. This is not the case: the percentage of aligned patients is 24.4\% in the heldout dataset and 27.7\% in the complete dataset. This percentage is much smaller than 100\%, and the difference between the percentages in heldout data vs complete dataset can be explained by the size differences between the two datasets. Hence, the complete dataset was used.

When performing binary alignment between \kmddss{} and the RCT assignments, 104 of the 376 patients became aligned. For the aligned and non-aligned patients, Table \ref{tab:binaryA_table} depicts the average of THI score -- as value, as improvement over the baseline and as percentage of improvement. It is evident that whenever the RCT agrees with the top-1 suggestion of \kmddss{}, the improvement is higher and the absolute score values are lower (lower scores are better).

\begin{table}[htb]
\begin{tabular}{lccc}
\hline
\textbf{Binary alignment} & \multicolumn{3}{c}{Average of THI score} \\
& value & improvement & improvement percentage \\
\hline
TRUE: 104 & 23.60 & 22.60 & 45.8\%
\\
FALSE: 272 & 38.91 & 10.09 & 19.6\%
\\
\hline
\end{tabular}
\caption{\protect{\label{tab:binaryA_table}}Average THI score and average THI improvement (as value and as percentage)for aligned and non-aligned patients under \emph{binary alignment}; the improvement among aligned patients is more than twice as large as for non-aligned ones}
\end{table}

\begin{figure}[htb]
\includegraphics[scale=0.45]{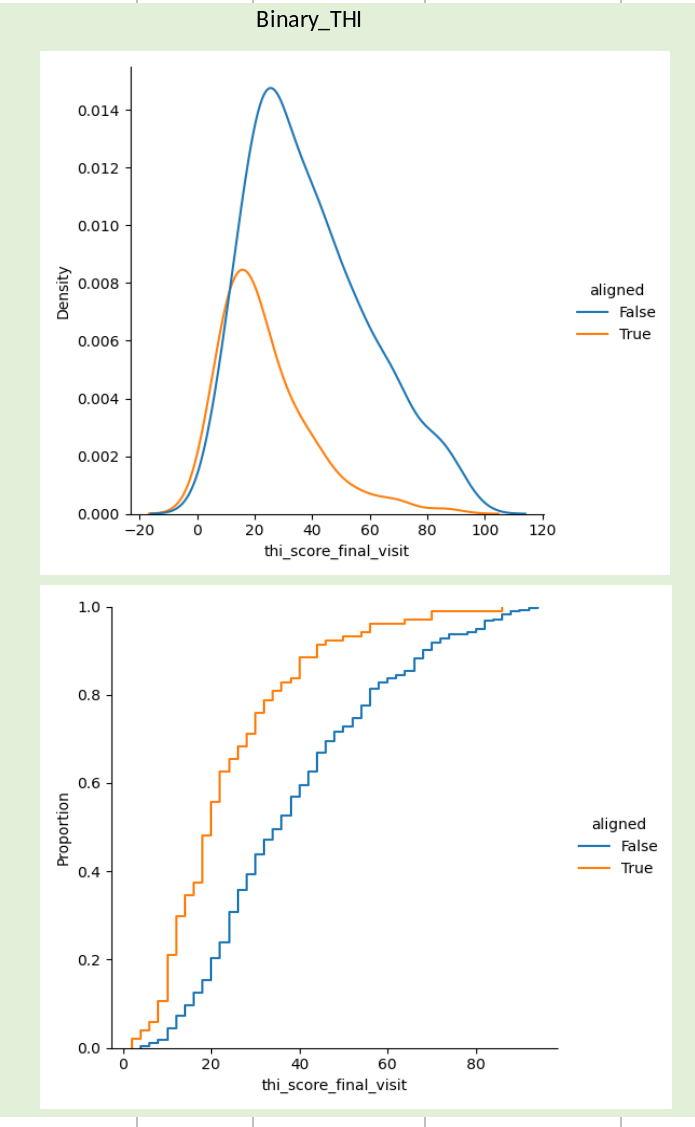}
\includegraphics[scale=0.45]{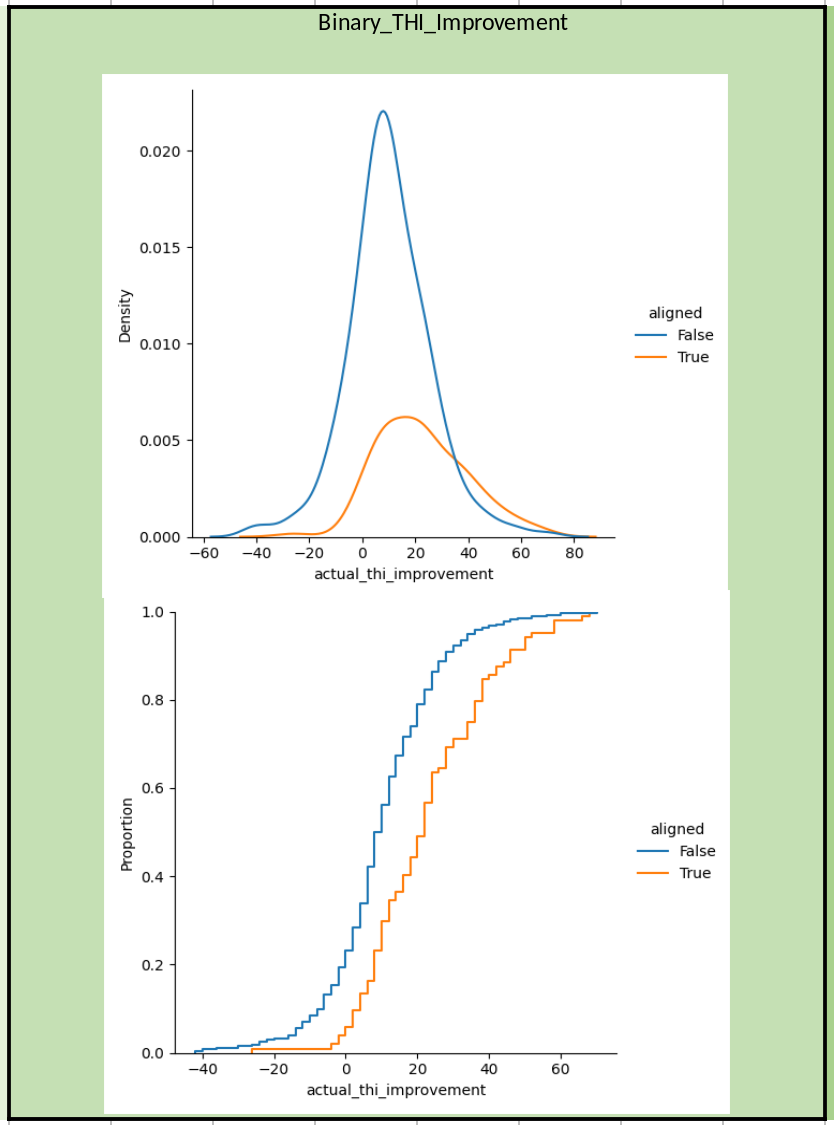}
\includegraphics[scale=0.45]{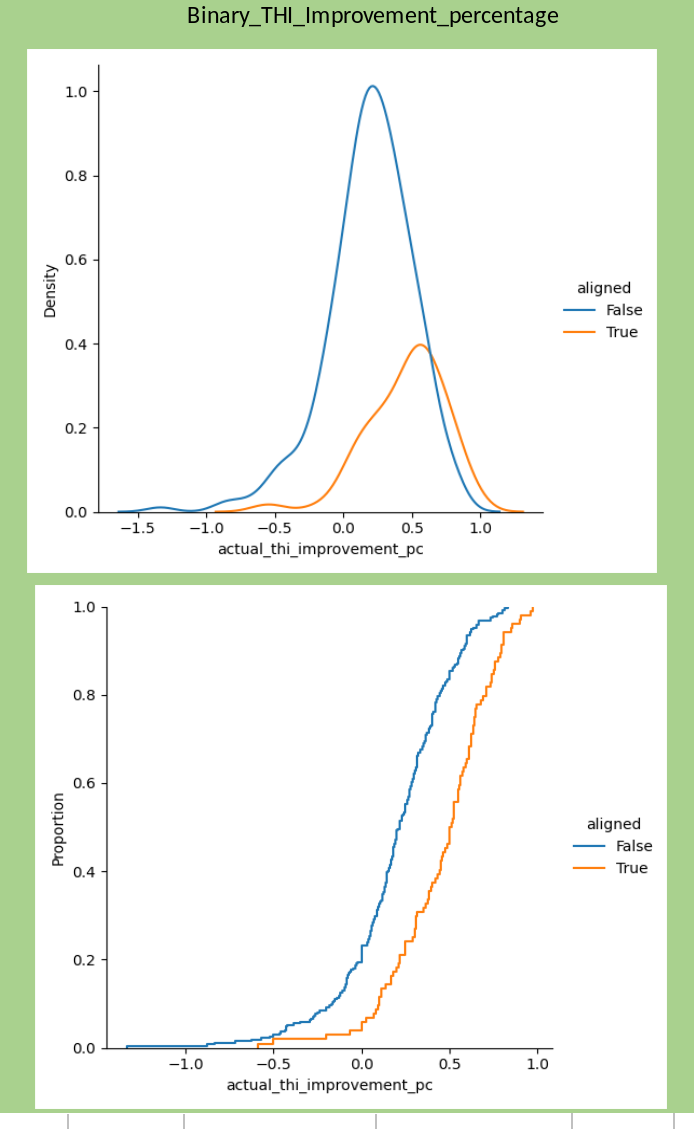}
\caption{\protect{\label{fig:binaryA_threestackedfigures}}Six plots on the THI score and its improvement (as value and as percentage) for aligned and non-aligned patients under binary alignment: the orange lines refer to aligned patients, the blue lines to non-aligned ones: all pairs of curves indicate that the distribution of scores, resp. improvements, is shifted more towards the 'better' numbers for the aligned patients than for the non-alined ones -- here 'better' refers to lower THI scores and larger improvement values, resp. improvement percentages}
\end{figure}

\paragraph{Visualizations} Figure \ref{fig:binaryA_threestackedfigures} contains 6 plots in three stacked figures. They depict the value distribution (upper subfigures) and percentage of population (lower subfigures) of THI scores (leftmost subfigures), THI improvement (middle subfigures) and percentage thereof (rightmost subfigures). The prefix 'Binary\_' refers to the two values (TRUE/FALSE) of the binary alignment. The orange curves refer to aligned cases (TRUE) and the blue ones to non-aligned cases (FALSE). It is noted that in the upper subfigures, the blue curve is always higher than the orange one, reflecting the fact that there are more non-aligned patients (more than 2.5 times as many) than aligned ones.

For the THI score, the orange density curve is more shifted to the left than the blue one, i.e. towards smaller THI scores (smaller is better). The orange population curve is also more shifted to the left, indicating that more of the aligned patients exhibit a smaller THi score than is the case for the non-aligned patients.

For the THI improvement and for the percentage thereof, the orange density curves are more shifted to the right than the blue curves, indicating that the improvement among the aligned patients is larger (larger is better). The orange population curves are accordingly shifted more to the right, towards the larger (better) values.

\paragraph{Statistical testing} We applied Welch's t-test to compare the means of aligned and non-aligned patients for THI score, THI improvement and percentage thereof. The results are shown on Table \ref{tab:binaryA_Welchttest}.

The null hypothesis that the means are the same is rejected in all cases. The p-values shown below are before Bonferroni correction, but obviously remain significant after correcting for multiple testing (three tests).

\begin{table}[htb]
\begin{tabular}{lrrr}
\hline
Means of THI score & statistic & p-value & df \\
\hline
value & -7.700 & 3.514e-13 & 240.742
% -7.700218741088722 & 3.514504693006503e-13 & 240.74219563493054
\\
improvement & 6.539 & 6.711e-10 & 172.835
%\textcolor{red}{6.53943635303847} & 6.711709850493907e-10 & 172.83573577255365
\\
improvement percentage & -8.453 & 2.998e-15 & 235.182
% -8.453264215467918 & 2.9984509776699217e-15 & 235.1820153285426
\\
\hline
\end{tabular}
\caption{\protect{\label{tab:binaryA_Welchttest}}Welch's t-test on the means of THI score, THI improvement and THI improvement percentage for aligned vs non-aligned patients: the p-values are before Bonferroni correction but remain significant after correcting for three tests -- all values are cut on 3 positions after the comma}
\end{table}

\paragraph{Focussing on patients with THI improvement of more than 7 units} Among the non-aligned patients, only 57.7\% experienced such an improvement, in contrast to 83.65\% of the aligned patients. We set the threshold to 7 units because this value is deemed 'significant' from the medical perspective \cite{zeman2011tinnitus}. %\todo{Citations on the 7 units; Vishnu proposes\\ \cite{zeman2011tinnitus} (in otherbib)}

\paragraph{Conclusion on binary alignment} These results indicate that when the treatment assigned to a patient by the RCT was the same as the one predicted by our \kmddss{}, then this treatment was also the best choice for the patients, and it has lead with higher likelihood (83.65\% vs 57.7\%) to 'significant improvement' of the THI score \footnote{The term 'significant' is not used in the statistical sense. It is borrowed from the medical terminology in the domain.}.

\subsection{Results for ternary alignment}
\label{subsec:ternaryA_Results}
For ternary alignment between \kmddss{} and RCT, Table \ref{tab:ternaryA_table} depicts the average of THI score -- as value, as improvement over the baseline and as percentage of improvement.The first row of full alignment is the same as for TRUE in Table \ref{tab:binaryA_table}. The second row depicts the 105 cases where \kmddss{} and RCT agree only on one of of a two-component treatment, so that there remain only 167 cases that are not aligned.  

\begin{table}[htb]
\begin{tabular}{lccc}
\hline
\textbf{Ternary alignment} & \multicolumn{3}{c}{Average of THI score} \\
& value & improvement & improvement percentage \\
\hline
2 (full alignment): 104 & 23.60 & 22.60 & 45.8\%
\\
1 (partial alignment): 105  & 34.04 & 14.0 & 28.2\%
\\
0 (no alignment): 167 & 41.98 & 7.63 & 14.26\%
\\
\hline
\end{tabular}
\caption{\protect{\label{tab:ternaryA_table}}Average THI score and average THI improvement (as value and as percentage)for aligned and non-aligned patients under \emph{binary alignment}; the improvement among aligned patients is more than twice as large as for non-aligned ones}
\end{table}

As can be seen on Table \ref{tab:ternaryA_table}, the average THI score is ca. 10 units lower for patients with full alignment than for patients with partial alignment; for non-aligned patients it is 7 units higher than for partially aligned ones. The differences in the improvement values are more drastic: the average improvement among partially aligned patients is twice as large as for non-aligned ones; for fully aligned patients, it is thrice as large.

\begin{figure}[htb]
\includegraphics[scale=0.45]{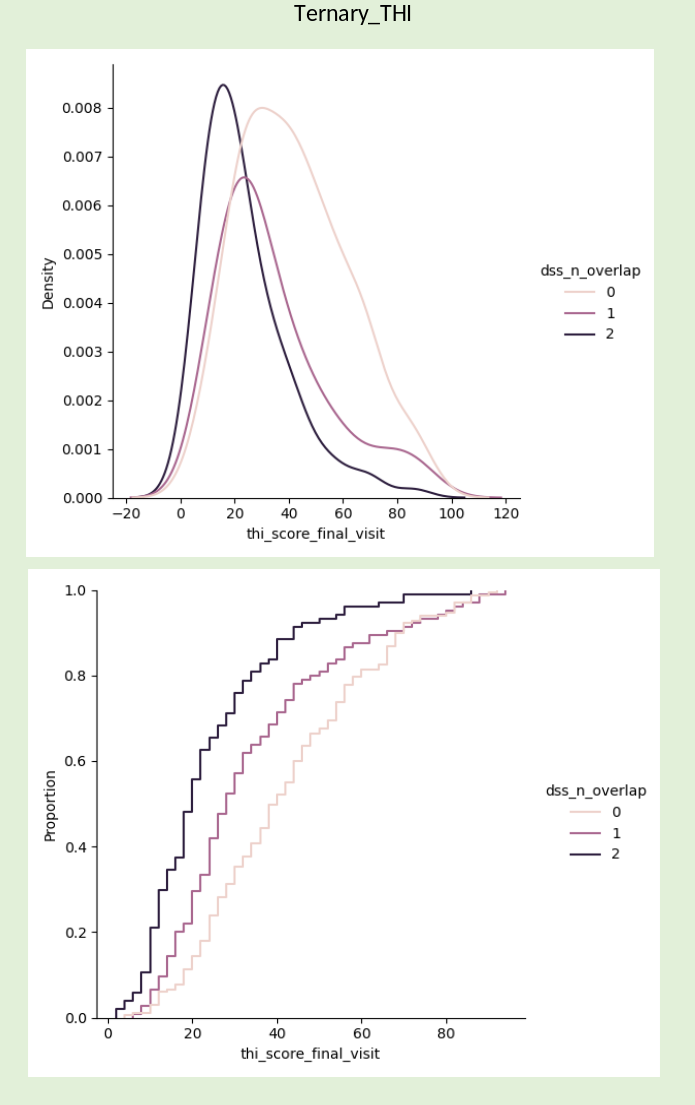}
\includegraphics[scale=0.45]{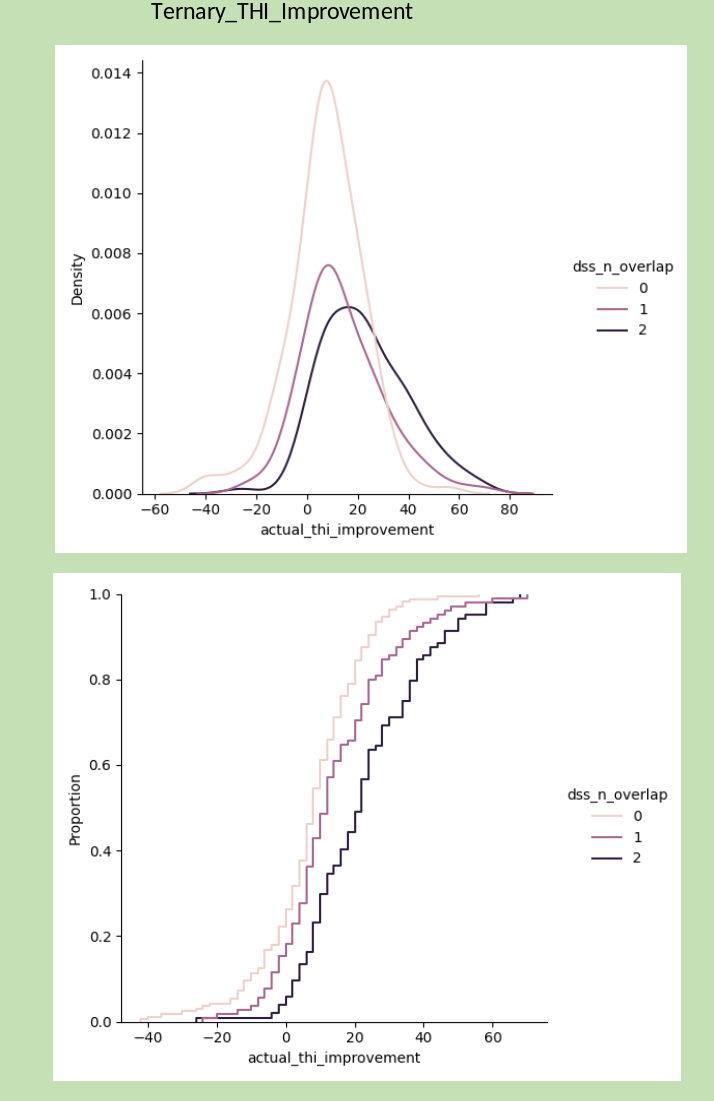}
\includegraphics[scale=0.45]{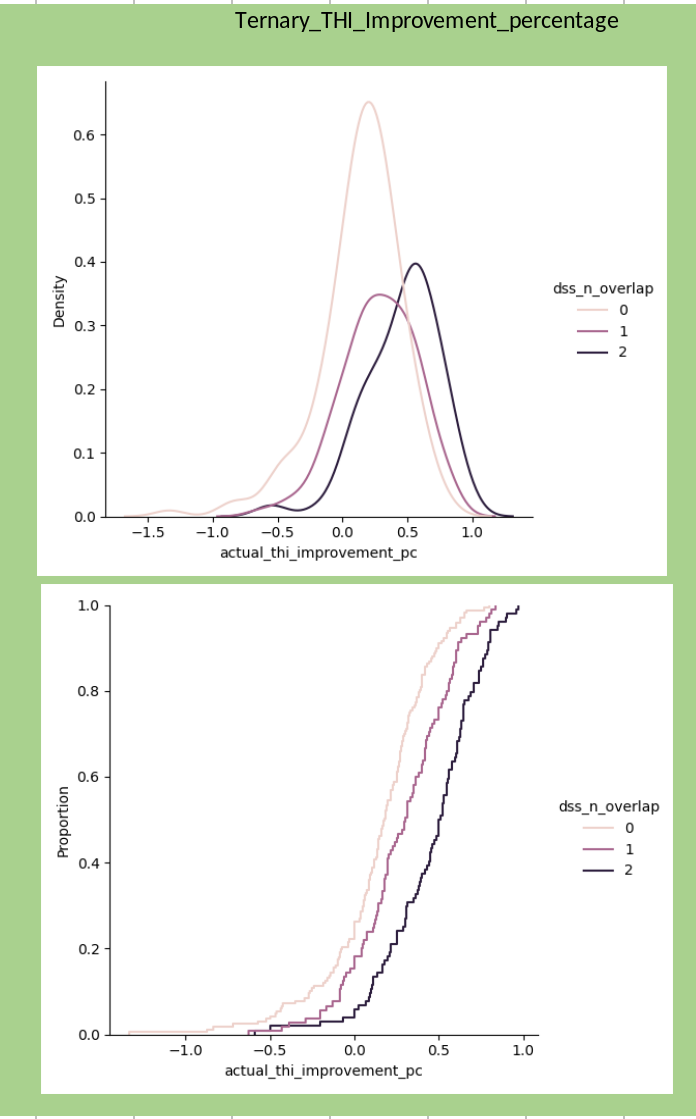}
\caption{\protect{\label{fig:ternaryA_threestackedfigures}}Plots on the THI score and its improvement (as value and as percentage) for fully aligned, partially aligned and non-aligned patients under ternary alignment: the grayblack lines refer to fully aligned patients, the dark pink lines to partially aligned ones, and the light pink lines to non-aligned ones: all plots indicate that the distribution of scores, resp. improvements, is shifted more towards the 'better' numbers for fully aligned patients, and less so for partially aligned ones, but still better than for the non-alined ones -- here 'better' refers to lower THI scores and larger improvement values, resp. improvement percentages}
\end{figure}

\paragraph{Visualisations} Figure \ref{fig:ternaryA_threestackedfigures} contains 6 plots in three stacked figures. They depict the value distribution (upper subfigures) and percentage of population (lower subfigures) of THI scores (leftmost subfigures), THI improvement (middle subfigures) and percentage thereof (rightmost subfigures). The prefix 'Ternary\_' refers to the three alignment values 0:none, 1:partial, 2: full alignment. The grayblack curves refer to fully aligned cases, the dark pink ones to partially aligned ones and the light pink cases to non-aligned ones. The tendencies observed under binary alignment are also observed here: the curves for the set of patients with full alignment are shifted towards the better values (smaller THI scores, larger THI improvements).

\paragraph{Focussing on patients with THI improvement of more than 7 units} Among the non-aligned patients, only 53.89\% experienced such an improvement, to 63.81\% among the patients with partial alignment, and in contrast to 83.65\% of the aligned patients. %\todo{Todo resolved by vishnu. All full aligned patients improve by 83.65\%. Numbers identical since full dss overlap numbers are same as binary alignment numbers.}
%We set the threshold to 7 units because this value is deemed 'significant' from the medical perspective \textcolor{red}{CITATIONS HERE}.

\paragraph{Conclusion on ternary alignment} These results indicate that when the treatment assigned to a patient by the RCT was fully aligned to the one predicted by \kmddss{}, then this treatment was also the best choice for the patients, and it has lead with higher likelihood to 'significant improvement' of the THI score. Patients with partial alignment also experienced more improvement than non-aligned patients though less than fully aligned patients.

%% file: 6_discussion.tex
We first discuss our results from the clinical perspective, and then from the perspective of DSS validation.

\subsection{Findings from the clinical perspective}
The evaluation of the \kmddss{} performance for tinnitus treatment recommendation must be seen in the context of validating the effectiveness of tinnitus treatments. In their systematic review, \cite{KikidisEtAl:RCTclinDSS:JCM2021} discuss a large number of RCTs that investigated the effectiveness of a variety of single-component treatments. Most of these studies found that the studied treatment had some positive effects. Tinnitus heterogeneity persists though, in the sense that the phenotypes who benefit from each treatment are still not well understood, so that treatment unification initiatives are needed \citep{SchleeEtAl:TowardsAUnification:2021}; the RCT used for our analysis is the first one in this direction. The effectiveness of the single-component and multi-component treatments is reported in \citep{SchoisswohlEtAl:medrxiv2024}, and our study delivers a first validation of the clinical decision support system \kmddss{} that predicts the outcome for each of these treatments for a given patient.

% We have shown that 

% \color{moreFGcolor}
% We discuss literature on (i) primary outcome of tinnitus, (ii) DSS for tinnitus, (iii) DSS for chronical disease administration and (iv) DSS on multi-component treatments.

% \todoin[color=moreBGcolor]{Help needed from Winny:
% \begin{itemize}
% \item Research on multi-component treatments for tinnitus
% \item DSS for tinnitus
% \item DSS for chronical disease administration: what diseases to consider here?
% \end{itemize}
% }

\color{black}
\subsection{Findings on validation procedures for choice among multiple treatments}
%Since we concentrate on treatments that have not been deployed yet, the evaluation of DSS performance in controlled trials, as studied by \cite{KwanEtAl:BMJ2020}, cannot be transferred to our \kmddss{}.
%
\paragraph{Guidelines}
\cite{VaseyEtAl:2022} elaborate on different guidelines for 'the early-stage clinical evaluation of decision support systems driven by artificial intelligence' and propose one themselves (called DECIDE-AI), and distinguish among 'Pre-clinical development', 'Offline validation' and online evaluation phases (cf. Figure 1 of their study). Our \kmddss{} validation can be seen as either 'In silico evaluation' or as 'Silent/shadow evaluation', and the closest protocol is TRIPOD-AI \citep{CollinsEtAl:BMJ2021}.

Taking the TRIPOD checklist as basis
\footnote{\texttt{www.tripod-statement.org/wp-content/uploads/2020/01/\newline Tripod-Checlist-Prediction-Model-Development.pdf}, accessed January 2, 2024}, our current study focussed on 'Methods' items 7 (Predictors), 8 (Sample size), 9 (Missing data) and 10 (Statistical analysis methods) and on the 'Results' and 'Limitations' (hereafter). All other items are covered by the RCT description itself \citep{UNITIrct:Trials2023}.

TRIPOD-AI concentrates on model performance, comparison to a 'reference strategy' is indirectly incorporated into the statistical analysis, but there are no guidelines on how to build such a reference strategy as part of the validation procedure.

\paragraph{Reference strategies as part of the validation procedure}
\cite{GrolleauEtAl:medrxiv2023} used RCT data to validate their approach on renal replacement therapy, and their validation procedure involves specifying a primary outcome and a reference strategy. The reference strategy is a treatment, so they can compare on the outcome. When there are multiple treatments, this needs to be generalized, as done in \citep{JafarEtAl:AIIM2024}.

\cite{JafarEtAl:AIIM2024} use two reference strategies (called 'run-to-run algorithms' A and B) from the literature. Their application area is monitoring of diabetes patients, hence the validation involves the simulation of patient behaviour over an 8 weeks period, thereby also specifying suboptimal diabetes administration scenaria. This would roughly correspond to suboptimal choices of treatment. On this basis, the recommendations of the DSS can be compared to the decisions of medical experts, who can decide between an optimal and a suboptimal treatment for a concrete case (i.e. patient).

This kind of validation has one caveat though: it demands that someone can distinguish between optimal and suboptimal options, i.e. inferior options (treatments) must be known in advance. This is rather difficult when the data come from an RCT that was itself intended to compare treatments. Accordingly, our validation procedure specifies the RCT assignment of treatments as reference strategy and, since this assignment is at random, it uses alignment between DSS and RCT as basis of comparisons.
%When the RCT is on the comparison of treatments, though, the assignment of the treatments or the ranking of the treatments must be incorporated into the reference strategy.
Hence, under the premise of using only the RCT data for the validation, our approach is unique and contributes to the body of knowledge on how to validate in such a restrictive scenario. Can we use data beyond the RCT for validation though?

\paragraph{Creating a matching dataset for validation}
%\cite{WheatonEtAl:2023}, \cite{ItzhakEtAl:AIIM2023}, \cite{ChenEtAl:2023}, \cite{McLernonEtAl:2023}, \cite{LeeEtAl:BMJ2022}
Studies that use RCTs to assess the performance of decision support tools assume that study participants are assigned to the DSS or to another procedure -- and thus come from the same population of patients handled by a given clinic. For early validation, this procedure is not feasible, so building up matching datasets seems a more appropriate procedure.

\cite{WheatonEtAl:2023} propose a method for matching single-arm RCTs with single-arm observational studies on similarity. Across a similar thread, \cite{ChenEtAl:2023} used matching to evaluate how well an outcome is predicted by clusters of symptoms: they selected questionnaires from two RCTs, and then applied hierarchical clustering with Manhattan distance to build clusters of patients and thus replicate their clusters of symptoms.

The approach could be transferred to the individual RCT arms of a multi-arm RCT, though only for treatments that have already been deployed. However, our earlier work \cite{PugaEtAl:FNINS2022} indicates that such a matching may lead to very small subsamples: when juxtaposing the observational data of two tinnitus centres
\footnote{Both centres participated in the RCT reported in \citep{UNITIrct:Trials2023}.}, we found that they used different, partially overlapping assessment batteries, that the distributions on age and gender were different, and that the values of the primary outcome at baseline (i.e. before treatment) were closer for the female patients than for the male patients. Besides, the implementations of the same treatment were slightly different (e.g. in duration). Hence, a reliable matching would demand the identification of individual patients that are as similar as possible to the data records used for training. For new treatments, as are the combinational treatments in our example RCT, such a matching would not be feasible at all.

Hence, building up a set of matched patient records as part of the validation procedure is a reasonable endeavour but seems limited to applications where the assessment batteries and the treatment specifications are fully standardized rather than new ones.

%\cite{VaseyEtAl:2022}, \cite{MathewsEtAl:NPJ2019}, \cite{NwanosikeEtAl:IJMI2022}

%\paragraph{(ii) Process mining}
% \cite{LeemansEtAl:AIIM2023}\cite{ChenEtAl:sysrev:AIIM2023}, \cite{MichalowskiEtAl:AIIM2023}, \textbf{\cite{CartolovniEtAl:IJMI2022}}

%\paragraph{Others - from Review papers}
%\textbf{\cite{SuttonEtAl:NPJ2020}}, \cite{TavazziEtAl:sysrev:AIIM2023}, \cite{LeiserEtAl:review:AIIM2023}, \textbf{\cite{BicaEtAl:2021}}, \textbf{\cite{RamgopalEtAl:2023}}, \textbf{\cite{JaspersEtAl:JAMIA2011}}

\color{black}

\subsection{Threats to validity}
\label{subsec:threats}
A first thread to validity is in the solution of the missing evidence pursued through the therapy-level models: when exploiting the data of multiple RCT arms that involve a given therapy, the corresponding therapy-level model indirectly takes evidence from other therapies into account. Were there training and test data of adequate size available per RCT arm, we could have compared the predictive performance of each therapy-level model to the performance of a model trained only on data of the RCT arm, and thus quantify the effects of mixing evidence from other arms. This will become possible after a deployment of all treatments at one or more of the clinical centres that were involved in the RCT. Nonetheless, our design ensures that a therapy-level model for therapy $T$ benefits from the fact that $T$ is present in multiple arms, while each of the other therapies is present in much fewer data -- 25\% in our RCT data.

A further thread to validity is in the design of the alignment-based validation procedure. The alignment assumes that a clinician would choose the top-ranked treatment of the \kmddss{}, and would ignore the confidence of the model. Albeit model confidence during treatment ranking is reasonable, incorporating this confidence in the validation procedure demands some caution: when predicting the outcome improvement per treatment for a patient, it is likely that the \kmddss{} is more confident in its prediction for the treatment that the patient actually received than for the other treatments, where the counterfactual treatment outcome was used instead.

Lastly, the usage of the complete RCT dataset for the alignment-based validation implies that some of the evidence used for training may have been leaked to the validation. Nonetheless, we found experimentally that the likelihood of alignment was comparable in the heldout dataset vs training dataset and did not exceed 30\% in either case, hence the \kmddss{} did not overfit.

%that the \kmddss{} did not overfit, since the likelihood of alignment was comparable in the heldout dataset vs training dataset and did not exceed 30\% in either case.

%% file: 7_conclusion.tex
%\color{MyraFGcolor}
%
%\subsection{Summary and Outlook}
%\label{subsec:outlook}
We have proposed \kmddss{}, an algorithm that predicts the outcome of each treatment for a list of \emph{new} treatments that may be offered to a patient. The treatment being new, \kmddss{} is designed as preparatory step for the deployment of treatments in a clinical setting where they have not been used yet. To train, test and validate our approach, we devised solutions that allow learning from the RCT data that compared the treatments. We addressed the issues of missing rationale in the treatmet assignment, of missing verification evidence when predicting the outcome for another treatment than the one actually assigned to the patient, and of missing evidence in general for some treatments/arms that have less patients than other arms. We devised a validation procedure that evaluates treatment outcome prediction and compares to the outcomes of the RCT after aligning it to the \kmddss{}. We have shown that our approach agrees with the early validation guidelines for DSS and is innovative in delivering an operationable solution despite learning and validating on the same RCT data.

The most urgent future task is the extension of the approach to take the confidence of the predictions into account and thus solve cases of similar predicted improvement but different confidence. Since the confidence is likely to be higher for the treatment actually delivered to the patient by the RCT, i.e. at random, the artefact can be best alleviated by establishing an independent dataset through patient matching.

Patient matching for new treatments is infeasible, hence we will investigate mechanisms that allow matching on the basis of similarity between a new treatment and existing components of this treatment. Finally, we will investigate ways of boosting the few RCT data through synthetic data generation, so as to alleviate the missing evidence problem.

%\begin{itemize}
%	\item generation of synthetic data
%	\item RCT in a clinic
%	\item ranking on the Pareto front
%\end{itemize}

\color{black}